\documentclass[letterpaper, 10 pt, conference]{ieeeconf}  

\IEEEoverridecommandlockouts                              
\usepackage[latin1]{inputenc} 
\usepackage{graphicx} 
\usepackage{epsfig} 
\usepackage{ifpdf}

\ifpdf
	\usepackage{epstopdf}
\else
\fi
\DeclareGraphicsExtensions{.eps,.pdf,.png,.jpg,.mps}

\usepackage{mathptmx} 
\usepackage{times} 
\usepackage{amsmath} 
\usepackage{amssymb}  
\usepackage[english]{babel}
\usepackage{url}
\usepackage{listings}
\usepackage{hyperref}
\usepackage{breakurl}
\usepackage{floatflt}
\usepackage{supertabular}

\title{\LARGE \bf
A Graphical Language for Real-Time Critical Robot Commands
}

\author{Andreas Angerer, Remi Smirra, Alwin Hoffmann, Andreas Schierl, Michael Vistein and Wolfgang Reif%
\thanks{The authors are with the 
	Institute for Software and Systems Engineering, 
	University of Augsburg, D-86135 Augsburg, Germany. 
	E-mail of corresponding author: {\tt angerer@informatik.uni-augsburg.de}
	}
\thanks{This work presents results of the research project \emph{SoftRobot} which is funded by the European Union and the Bavarian government within the \emph{High-Tech-Offensive Bayern}. The project is carried out together with KUKA Labs GmbH and MRK-Systeme GmbH and is kindly supported by \mbox{VDI/VDE-IT GmbH}.}
}

\begin{document}

\maketitle
\thispagestyle{empty}
\pagestyle{empty}

\begin{abstract}
Industrial robotics is characterized by sophisticated mechanical components and highly-developed real-time control
algorithms. However, the efficient use of robotic systems is very much limited by existing proprietary programming methods. In the research project SoftRobot, a software architecture was developed that enables the programming of complex real-time critical robot tasks with an object-oriented general purpose language. On top of this architecture, a graphical language was developed to ease the specification of complex robot commands, which can then be used as part of robot application workflows. This paper gives an overview about the design and implementation of this graphical language and illustrates its usefulness with some examples.
\end{abstract}


\section{Introduction}
\label{sec:introduction}

\begin{figure}[thbp]
	\centering
		\includegraphics[width=\linewidth]{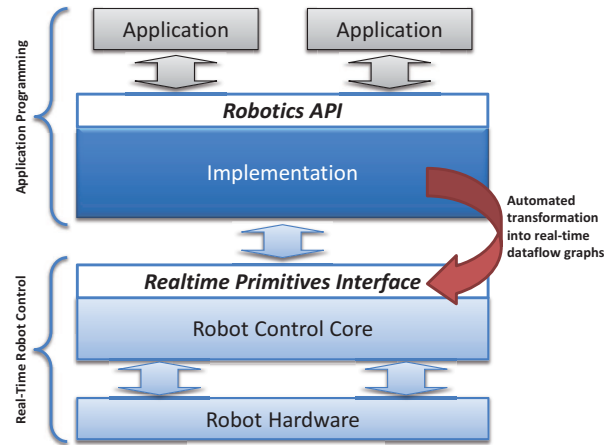}
	\caption{Robot application are programmed against the \emph{Robotics~API}. High-level commands specified using the Robotics API are automatically transformed into dataflow graphs at runtime and executed with real-time guarantees.}
	\label{fig:Architecture}
\end{figure}

When programming robots to perform long-lasting and complex tasks, the programmer usually wants to abstract from the technical details of controlling the robot hardware, e.g.\ hard real-time constraints, closed-loop controllers, or controller parameters. The focus should rather lie on the \emph{what} aspect of the task~\cite{Pires2009}. In the research project \textit{SoftRobot}, an extensible software architecture~\cite{Hoffmann2009} has been developed to both facilitate the development of robotic applications with the aforementioned level of abstraction, and at the same time keep real-time constraints in mind. This multi-layer architecture (cf.\ Fig.~\ref{fig:Architecture}) allows to program industrial robots using a standard, high-level programming language (e.g.\ Java) and, at the same time, ensures that commands are executed on the robot hardware with real-time guarantees.

The SoftRobot architecture provides application programmers with the Java-based \emph{Robotics API}~\cite{Angerer2010}, an object-oriented, extensible application programming interface for robot applications. The Robotics API contains an open domain model of (industrial) robotics describing the available actuators and devices, as well as possible actions and tasks, and also includes ways of maintaining a world model of the relevant parts of the environment. The Robotics API allows to specify actions to be executed by actuators (for example, a linear movement by a robot), and to combine multiple actions in different ways to create more complex commands. Every such command is guaranteed to be executed with certain real-time guarantees considering the timing of all actions and their combination. As an example, consider the case of a line welding application in the automotive industry: robots equipped with welding tools perform welding operations following series of continuous, complex geometric shapes. The workpieces in this case are (quite expensive) car bodies, which requires sophisticated handling of process errors in order to not damage the bodies. If an error occurs during a welding operation, the welding tool has to be turned off immediately (meaning with a maximum delay of at most some milliseconds) and probably even needs to be moved away quickly from the surface. Such operations have to be executed with hard real-time guarantees, while application developers still want to be able to specify them on a high level of abstraction and in an application-specific way.

To meet those real-time requirements, the Robotics API relies on the \emph{Robot Control Core} (RCC) layer. At runtime, Robotics API commands are transformed to a data-flow language called \emph{Realtime Primitives Interface} (RPI)~\cite{Vistein2010} which is implemented by the RCC. The dataflow language consists of (robotics-specific) calculation blocks which are referred to as (real-time) \emph{primitives} and are connected by data-flow links to form a graph, referred to as \emph{primitive net}. During execution of a primitive net, each primitive is evaluated cyclically. The primitives have known worst-case execution time and thus allow the execution of the task in a deterministic manner. As the RCC is responsible for real-time execution of primitive nets and for controlling the robotic hardware, it must be running on a real-time operating system. A reference implementation of an RPI-compatible RCC~\cite{Vistein2010} was developed using OROCOS~\cite{Bruyninckx2001} and Linux with Xenomai real-time extensions.

Robotics API commands and their combination can be specified using mechanisms of the object-oriented Java language, such as instantiation of command objects and composition of commands by calling respective API methods. This mechanism proved to be very flexible and able to express a large variety of complex tasks that are common in the industrial robotics domain. However, practical experience has also shown that the definition of complex command structures often results in long and confusing program code. To overcome this, several approaches have been considered and evaluated. Some efforts have been made to introduce a simpler API while at the same time still providing enough flexibility and extensibility. Another idea that was investigated and constitutes the main contribution of this work is the development of a graphical formalism for specifying such command structures, as they seemed quite well suited for being expressed graphically. The resulting graphical specification tool proved to be a real step forward and enables much quicker specification of Robotics API command structures. Code generated from such a specification can be easily used inside Robotics API programs. In this paper, we present the design and implementation of this graphical language and illustrate its usefulness with examples.

Before presenting the main contribution, the next section will give an overview about existing work in the area of graphical programming languages for robots. The following Section~\ref{sec:rapi} describes the characteristics of the Robotics API and its command model. Sect.~\ref{sec:language} describes the design principles of the proposed grapical language, whereas Sect.~\ref{sec:realization} presents details of its realization. Sect.~\ref{sec:examples} presents a practical example of a robot task specified with the developed language. Finally, Sect.~\ref{sec:conclusion} gives a conclusion and an outlook.

\section{Related Work}
\label{sec:related}

Roboticists have developed various tools for graphical specification of robot actions. Some frameworks relying purely on dataflow-based specification of robot actions like ControlShell or ORCCAD provide graphical editors for their dataflow diagrams~\cite{Schneider1998}\cite{Borrelly1998}. In~\cite{Bredenfeld2001}, graphs of data processing operators based on the Dual Dynamics architecture can be specified graphically in order to synthesize robot controllers. 

While the aforementioned approaches are roughly related to the RPI layer in the SoftRobot architecture, other approaches operate on higher abstraction levels. MissionLab supports graphically specifying configurations of inter-related robot behaviors~\cite{MacKenzie1997}. Manufacturers of industrial robot systems like KUKA and ABB are also investigating the possibilities of graphical languages for robotics and are proposing approaches~\cite{Bischoff2002}~\cite{Chen2009}.

Other frameworks used in the robotics research community today also incorporate graphical tools. The Robotics Developer Studio~(RDS) from Microsoft includes the Visual Programming Language~(VPL)~\cite{Morgan2008} that can orchestrate services created with the RDS. The research robot NAO by Aldebaran Robotics comes bundled with the development tool Choreographe that also includes means for graphically orchestrating robot behaviors~\cite{Pot2009}. State Machines are also often used for implementing robot behavior. The ROS\footnote{http://www.ros.org} framework includes at least a visualization tool for State Machines implemented using the tool SMACH\footnote{http://www.ros.org/wiki/smach}.

The Command model used in the RoboticsAPI was designed to introduce a level of abstraction that is useful for programming complex tasks of industrial robotic systems (for details see~\cite{Hoffmann2010a}). At the same time, those Commands are designed to be transformable to real-time primitives nets that the Robot Control Core can execute, guaranteeing hard timing constraints. These different kinds of constraints led to some special characteristics of the designed Command model that neither make it a data-flow language, nor directly comparable to existing paradigms like Software Services or Abstract State Machines. Thus, reusing parts of existing graphical languages seemed to be infeasible. For this reason we propose a new Domain Specific Language tailored to the RoboticsAPI Command Model. The next sections will describe the structure of the RoboticsAPI more in detail and after that present the proposed graphical language for specifying RoboticsAPI Commands.

\section{Structure of the RoboticsAPI core}
\label{sec:rapi}

\begin{figure}[thbp]
	\centering
		\includegraphics[width=\linewidth]{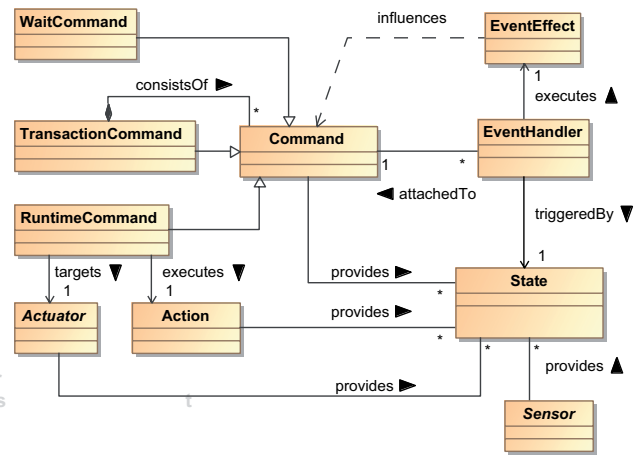}
	\caption{Class structure of the RoboticsAPI core}
	\label{fig:rapioverview}
\end{figure}

The RoboticsAPI is intended to serve as an extensible object-oriented framework for creating robotic applications. Due to its framework nature, its core parts define a general class structure that may be extended for use with concrete hardware and creating specific applications. In this section, we give an overview of this core structure. Fig.~\ref{fig:rapioverview} depicts the most basic concepts and their relationships. In the following, we will explain all parts step by step.

Hardware devices that can be controlled in any way are modeled as \emph{Actuators} (cf.~Fig.~\ref{fig:rapioverview}, bottom left). An example of a more specialized Actuator would be a robot arm. Actuators have some properties (e.g. the number of joints) and can be configured in certain ways (e.g. defining the maximum allowed velocity of motions). Actuators in the RoboticsAPI in general do not contain code that implements the execution of any kind of action (e.g. the interpolation of motions). They can rather be seen as proxy objects representing devices in the Robot Control Core (cf.~Fig.~\ref{fig:Architecture}).

Actions that Actuators should execute are modelled by the class \emph{Action}. An example would be a linear motion of a robot arm in space. Like Actuators, Actions carry parameters (e.g. the velocity of a linear motion). To let an Actuator execute an Action, a \emph{RuntimeCommand} has to be defined. It combines exactly one Actuator with exactly one Action. Like all other \emph{Commands}, RuntimeCommands provide methods to actually execute them. When one of those methods is called, the specification of the Command is transformed into a primitive net, which is then sent to the RCC and executed there in a real-time environment. Details on this transformation and execution process can be found in~\cite{Schierl2012a}.

To create more complex task specifications (e.g. a robot motion during which a tool action should be executed), the other types of \emph{Commands} can be used. A \emph{TransactionCommand} is a composition of other Commands. All contained Commands (and their scheduling rules, see below) are executed in the same real-time context. This means that, upon execution, they are transformed into different parts of the same primitive net and executed atomically by the RCC. To define scheduling rules, \emph{EventHandlers} can be used. EventHandlers are attached to Commands and can influence their execution by executing \emph{EventEffects}. An example would be stopping a Command when certain events occur. When attaching an EventHandler to a  TransactionCommand, it is e.g. also possible to let the EventHandler start one of the child Commands contained in the TransactionCommand.

EventHandlers can be triggered upon changes in the state of the system. The concept \emph{State} models a certain part of this system state. A State can be active at any point in time, or not active. An EventHandler can be triggered when the activeness changes, i.e. when the State is becoming active or inactive. States can be provided by Commands as well as Actuators, Actions and \emph{Sensors}. Commands provide States expressing e.g. that the Command has been started or stopped. Actuators provide mostly error States (e.g., a robot's drives are disabled), while Actions can provide States like e.g. a Motion having been executed to a certain degree. Sensors are providing data that is measured in some way. Boolean-type sensors (like a digital field bus input measuring a high or low value) provide a State that directly corresponds to the measured value. For more complex sensor data, the RoboticsAPI allows defining derived Sensors that process this data in order to provide States (e.g. the condition that 'a laser scanner sees an object in a distance smaller than a value X' can be expressed as a State).

All the concepts presented above can be used to create arbitrarily complex structures of Commands. Due to the fact that each such structure is guaranteed to be executed with hard real-time timing guarantees, a large variety of applications can be realized. The RoboticsAPI was already used to e.g. perform robot motions with collision detection, for force-based part assembly with defined maximum force or for cooperative transport of workpieces by multiple robot arms, requiring tight synchronization of the arms.

To illustrate the composition of Robotics API commands, Listing~\ref{lst:commandexample} gives a Java code example. Here, a robot is instructed to execute a simple Point-To-Point (PTP) motion in joint space, while at a certain progress of the motion, its gripper is opened by setting a field bus output. This simple example shows that the code required to define a hierarchy of Commands and rules for scheduling their execution tends to get long and complex. Worse than the pure code size is the fact that it is hard to read the ``big picture'' (i.e. the resulting Command) from this kind of definition, as the hierarchical structure of Commands is hard to capture from this kind of textual definition. As stated previously, different ways to tackle this problem have been examined, with the approach presented in this paper being one of them.

\begin{lstlisting}[float,label=lst:commandexample,caption=Robotics API Command Example,emph={getSingle},emphstyle=\textit]

// RoboticsRuntime is the Factory for Commands
RoboticsRuntime rt = RoboticsRegistry.getSingle(
		"orocos", RoboticsRuntime.class);

// create motion RuntimeCommand for LWR
LWR lwr = RoboticsRegistry.getSingle("lwr", LWR.class);
PTP ptp = new PTP(
		lwr.getHomePosition(), 										// start 
		new double[] {1.57, 0, 0, 1.57, 0, 0, 0} 	// goal
		);
RuntimeCommand ptpCmd = rt.createRuntimeCommand(lwr, ptp);

// create command for setting field bus output
DigitalOutput o = RoboticsRegistry.getSingle(
		"gripperClose", DigitalOutput.class);
SetDigitalValue close = new SetDigitalValue(true);
RuntimeCommand closeCmd = rt.createRuntimeCommand(o, set);

// combine both commands
TransactionCommand trans = rt.createTransactionCommand();
trans.addStartCommand(ptpCmd); // auto-start Command
trans.addCommand(closeCmd);		 // no auto-start

trans.addStateEnteredHandler(
		ptp.getMotionTimePercent(30), // State
		new CommandStarter(closeRt)); // EventEffect

// execute TransactionCommand
trans.execute();
\end{lstlisting}

\section{Design of the Graphical Language}
\label{sec:language}

In this section, we present the design of the GSRAPID language. GSRAPID stands for \textbf{G}raphical \textbf{S}oftRobot \textbf{R}obotics\textbf{API} \textbf{D}iagram and refers to the graphical models that can be defined. In the following, all important concepts of this language are explained.

\subsection{Basic Operators}
\label{subsec:basicoperators}


The graphical language designed to specify complex RoboticsAPI Commands contains a set of basic graphical operators. Those operators represent mainly the concepts presented in the previous section. Furthermore, some operators allow to define relationships between other operators to resemble relationships between Commands. 

There are two ways of defining a relationship between entities: nesting and connecting.
If an object A (e.g. a State) is \emph{nested} inside another object B (e.g. a Command), then object A is graphically surrounded by object B. This relationship doesn't always fully specify the interaction between the nested objects and the surrounding ones, but defines a cohesiveness that can carry semantics.
The scheduling of different Commands is achieved by \emph{connecting} different graphical objects which results in a graph, resembling a complex RoboticsAPI Command.

\begin{table}
\bottomcaption{Basic graphical operators}
\label{tab:basicoperators}
\begin{supertabular}{@{}p{0.22\columnwidth}@{}p{0.78\columnwidth}@{}}
\hline
\vspace{2mm}
\includegraphics[scale=.16]{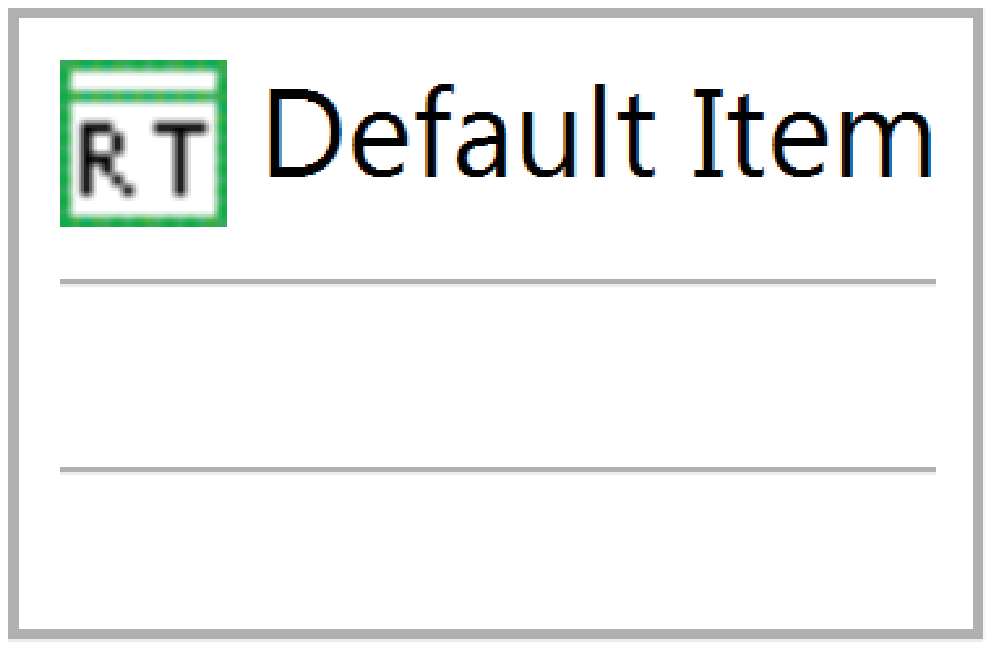} & A nestable item like this can either represent any type of Command, an Actuator, an Action, or a Sensor, depending on the icon in the upper left corner. It supplys one compartment for the inner nestable objects and one compartment for inner States.\\
\hline
\vspace{0.1mm}
\includegraphics[scale=.18]{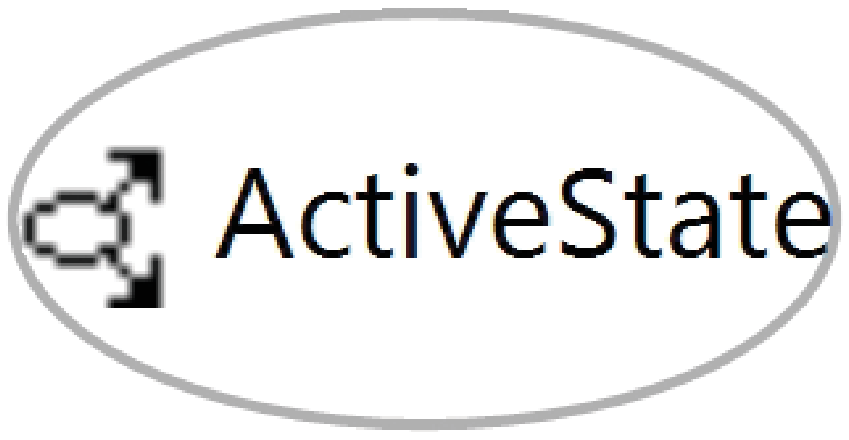} & Regular State item, which belongs to a Command. The label identifies the type of State.
\\
\hline
\vspace{2mm}
\includegraphics[scale=.2]{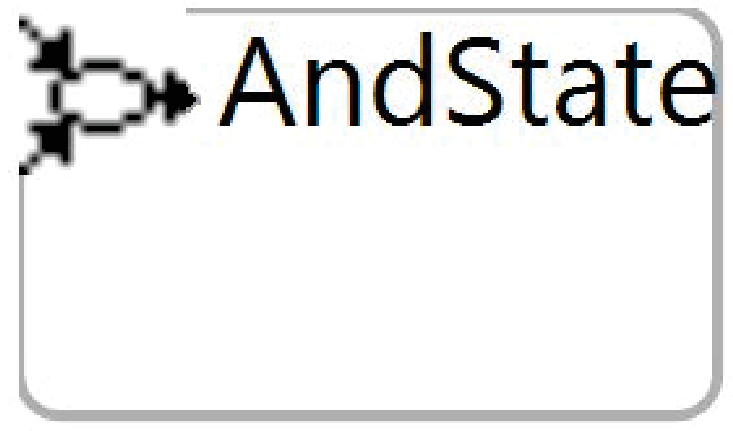} & LogicalState, which is composed of other States by ingoing connections. The rectangular shape suggests the composable character of this item, while the round corners identify it as a State.\\
\hline
\vspace{0.1mm}
\hspace{1mm}
\includegraphics[scale=.22]{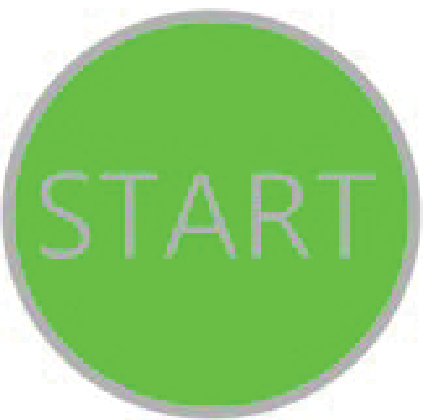} & Represents an entry-point for a Diagram from where the contained Commands can be started.\\
\hline
\vspace{0.1mm}
\includegraphics[scale=.37]{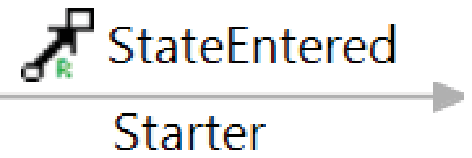} & Connection between a State and a Command. The labels express the event handler parametrization described in section~\ref{subsec:commandscheduling}.\\
\hline
\vspace{4mm}
\includegraphics[scale=.7]{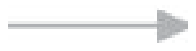} & Connection either between a State and a LogicalState, or between a starter and a Command. There is no labeling needed, since the context of usage clearly defines its function.\\
\hline
\end{supertabular}
\end{table}

The basic graphical operators are defined in Table~\ref{tab:basicoperators}. In addition to the pure graphical definition, each operator may have a set of properties attached. The values of those properties can be specified separately from the language diagram in a dedicated property editor. By this separation of graphical representation and property values, the visual clarity of the language is preserved.

\subsection{Diagrams and Commands}
\label{subsec:diagramscommands}

\begin{figure}[thbp]
	\centering
		\includegraphics[width=\linewidth]{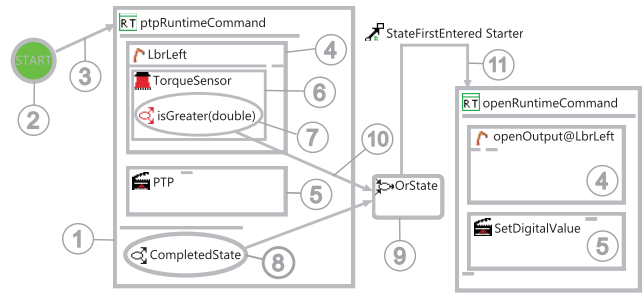}
	\caption{A Diagram defined in the graphical language}
	\label{fig:concept_screen_canvas}
\end{figure}

Graphical RoboticsAPI Commands are specified inside a so called \emph{Diagram}. A Diagram is considered the top-level of a graphical Command specification. During code creation, it is translated to a Robotics API TransactionCommand. To make a diagram a valid Command specification, the following rules need to be followed:
\begin{itemize}
	\item At least one Command (cf. Fig. \ref{fig:concept_screen_canvas}, item marked 1) has to be present within the created Diagram.
	\item There has to be at least one entry-point for the command (cf. Fig. \ref{fig:concept_screen_canvas}, 2)
\end{itemize}

In order to start a command from one of the entry points, so called "StarterConnections" (cf. Fig. \ref{fig:concept_screen_canvas}, 3) have to point to the commands that shall be started first. Further Commands can be added to Diagrams as needed. TransactionCommands inside Diagrams are graphically modeled as special nestable items. They can themselves contain nestable items, in this case further Commands (TransactionCommands or RuntimeCommands).

RuntimeCommands are also modeled as nestable items. In contrast to TransactionCommands, they can only contain exactly one Action and one Actuator as inner items. As described in \ref{sec:rapi}, this specifies that the Action has to be executed by the Actuator. An example is shown in Fig.~\ref{fig:concept_screen_canvas}. The element labeled (1) is a RuntimeCommand with an Actuator (4) and an Action (5) nested inside.

\subsection{Sensors and States}
\label{subsec:sensorsstates}

The prerequisite for defining execution-time scheduling dependencies between RoboticsAPI Commands are States (see Sec.~\ref{sec:rapi}). As explained before, States can be provided by Commands, Actuators, Actions or Sensors. To reflect this in the graphical language, States can be nested inside the graphical operators that resemble those concepts. 

Sensors themselves can be embedded in a Diagram in two ways: They can either be nested in an Actuator (cf. Fig. \ref{fig:concept_screen_canvas}, 6), or, depending on the type of Sensor, be defined as a toplevel sensor in the Diagram.
As States can be defined based on Sensor data (e.g. the State that a force sensor measures a value greater than 5N), graphical State items can be nested inside Sensor items (cf. Fig. \ref{fig:concept_screen_canvas}, 7).
States, however, are not limited to Sensors and can also be added to the graphical representations of Actions, Actuators and Commands. Fig. \ref{fig:concept_screen_canvas} shows a CommandState (8) which belongs to the RuntimeCommand and represents the state that this command has been completed. 


In addition to the regular States, LogicalStates have a special graphical representation (cf. Fig. \ref{fig:concept_screen_canvas}, 9). They are States that are derived from one or more other State(s). 
This deriviation is symbolized by the StateConnections in Fig. \ref{fig:concept_screen_canvas}, 10.
Like the other States, LogicalStates are connected to the commands they shall have an effect on by EventEffect-connections (cf. Fig. \ref{fig:concept_screen_canvas}, 11).

\subsection{Command scheduling}
\label{subsec:commandscheduling}
The event mechanism is the core concept for scheduling Commands in the graphical language. An EventHandler can be specified graphically by inserting an EventEffect connection originating from a State and targeting a Command. Further details concerning the scheduling are specified as properties of this connection and visualized as labels of the connection. These details include constraints specifying on which kind of event to react (State (first) entered/State (first) left) as well as the effect of the event. Possible effects are e.g. starting, stopping and cancelling the targeted Command.

With this kept in mind, the schedule in Fig. \ref{fig:concept_screen_canvas} could be expressed as:
"If the TorqueReachedSensor state has appeared for the first time on the Actuator LbrLeft or the RuntimeCommand \textit{ptpRT} is in a completed state for the first time, start the RuntimeCommand \textit{open}"

\section{Realization}
\label{sec:realization}

To enable users to create RoboticsAPI Commands with GSRAPID, a graphical tool called \emph{GSRAPID IDE} was created. The Eclipse IDE was chosen as the basis for this tool mainly for two reasons: On the one hand, there exists a variety of stable Eclipse-based frameworks for creating custom (graphical) languages (cf.~Sec.~\ref{subsec:technologies}). On the other hand, there is ongoing work on an Eclipse plugin that supports development of robot applications with the Robotics API. This plugin (called the RoboticsAPI plugin) provides some mechanisms that were useful for connecting the GSRAPID IDE to the environment of a robotics application, like e.g. the available robots in this application. We will not go into details about the RoboticsAPI plugin in this paper, but rather explain those mechanisms that were used in the realization of the GSRAPID IDE.

\begin{figure}[thbp]
	\centering
		\includegraphics[width=0.85\linewidth]{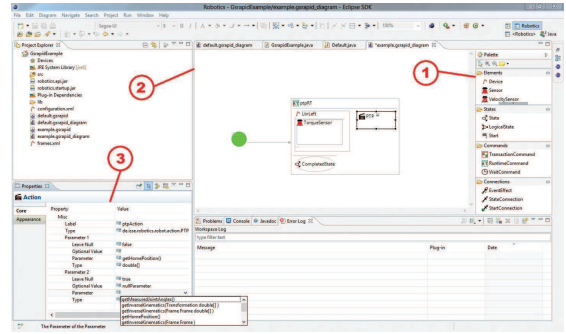}
	\caption{Plugin user interface for creating commands in the graphical language}
	\label{fig:plugin}
\end{figure}

The GSRAPID IDE basically consists of three important components (cf. Fig. \ref{fig:plugin}):

\begin{enumerate}
	\item The Tools palette, from which basic GSRAPID operators are selected
	\item A working Canvas, where GSRAPID Diagrams are created
	\item A Properties View, where specific attributes and parameters can be set for the currently selected graphical entity.
\end{enumerate}

The common working flow is dragging and dropping an element from the Tools palette onto the Canvas (to the right hierarchy level inside the Diagram) and then setting its properties in the Properties View.

\subsection{Technologies}
\label{subsec:technologies}
A variety of technologies was used to develop the GSRAPID IDE with all its components. At this point, we briefly present the concepts behind those technologies before their application in this context is described in the later sections.

\subsubsection{Eclipse Modeling Framework}
The \textit{Eclipse Modeling Framework} (EMF) allows to generate a complete Eclipse plugin based on abstract model definitions.
The advantage of using EMF for plugin development is, that a) a lot of code required for managing model instances at program runtime can be generated and b) the generated code is guaranteed to work correctly, according to the model definitions. 
EMF also provides support for creating a properties view, which is frequently used for defining GSRAPID Diagrams.

\subsubsection{Graphical Modeling Framework}
In order to create a graphical modeling environment, though, EMF is not sufficient.
This is where the graphical modeling framework (GMF) comes into place. 
It enhances the functionality of EMF by using and adapting components of Eclipse's graphical editing framework (GEF).
By defining three additional models, i.e. the graphical model, the tooling model and the mapping model, a stub for a completely functional graphical modeling plugin can be generated.

\subsubsection{Java Remote Method Invocation}
The Java Remote Method Invocation (RMI) enables the access to objects and methods executed in a different process or even on a different physical system. This mechanism is needed for some parts of the communication to the RoboticsAPI plugin, as explained later.
By design of RMI, the calls to functions hosted in a remote Java environment are almost identical to local function calls, resulting in a high degree of transparency. 
The only difference of remote calls is the necessity of additional exception handling, related to the dynamics of distributed environments~\cite{rmi_transparency}

\subsubsection{Java AST/DOM}
Java AST/DOM is one of Eclipse's most basic frameworks which is responsible for a lot of functionalities in the IDE, e.g. refactoring.
Each Java class definition is represented in Eclipse as an Abstract Syntax Tree (AST), which is similar to the Document Object Model (DOM), and is continuously synchronized.
This way, changes to the code always have an abstract representation which e.g. also makes code outlines available~\cite{ast}.

\subsection{Data Model}
\label{subsec:datamodel}
For gathering all data required for a fully specified GSRAPID diagram, some concepts were introduced in addition to the classes provided by the Robotics API.
The concepts of LogicalStates and the implicit Eventhandler were already discussed in \ref{subsec:sensorsstates} and \ref{subsec:commandscheduling}.
Additionally, a completely new \emph{Parameter} concept had to be introduced. It reflects parameters of graphical entities in order to meet the requirements for successful code generation.


This new Parameter object is needed for the correct initialization of states, sensors and actions. They usually depend on a set of input parameters, like e.g. start and goal positions of a robot movement Action. These parameters are specific to the respective Action and thus handled by the Property View of the GSRAPID IDE in a generic way.

The values of those Parameters can be constant values entered by the user\footnote{In case the parameters are primitive types like Integer, Boolean, etc.}. In some cases though, it is desirable to determine the parameter value dynamically by e.g. calling a method. This is supported by the dynamic method search capabilities of the GSRAPID IDE (cf.~Sec.~\ref{subsec:dynamicmethodsearch}). For all other cases (where neither a constant value nor a simple method call is sufficient), a parameter may be assigned a variable name. The code generator then creates setter methods for all those variables, which have to be called before the generated Command is executable.

\subsection{Dynamic Method Search}
\label{subsec:dynamicmethodsearch}
When specifying that a parameter of a GSRAPID operator should be determined by a method call at runtime, the object instances available at runtime have to be known, as well as their methods. The RoboticsAPI plugin maintains a special runtime context for each robotic application that is created as project in an Eclipse-Workspace. In this runtime context, an instance of the RoboticsAPI library is loaded and configured according to special configuration files of the respective Eclipse project. That means that in this context, e.g. instances of all RoboticsAPI Devices are created and can be accessed, as well as all class definitions inside the Robotics API. By using the mechanisms of Java Reflection\footnote{http://java.sun.com/developer/technicalArticles/ALT/Reflection/}, this runtime context can be browsed and methods can be searched that comply to certain requirements (e.g. returning values of certain datatypes).

The GSRAPID IDE employs these mechanisms and uses a heuristical approach to find methods that might be feasible for determining values of GSRAPID operator Parameters at runtime. The approach looks for appropriate methods in a certain ``context'' of the current GSRAPID diagram, which includes e.g. the Actuators and Sensors used, as well as the set of methods available in the Command classes.

\begin{figure}[thbp]
	\centering
		\includegraphics[width=0.85\linewidth]{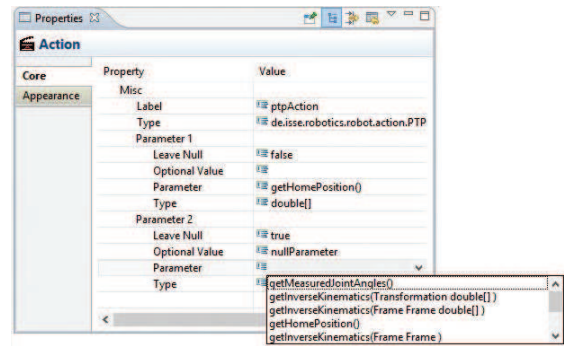}
	\caption{Setting the needed input parameters for an action}
	\label{fig:parameter_handling}
\end{figure}

As the runtime context for Eclipse projects is hosted as a separate operating system process by the RoboticsAPI plugin, Java RMI is used for communicating with those processes. Feasible methods are suggested to the user in a dropdown list in the properties (cf. Fig. \ref{fig:parameter_handling}). The GSRAPID IDE also supports recursive usage of this mechanism for determining arguments of methods using the same dynamic definition.

\subsection{Code Generation}
\label{subsec:codegeneration}
\begin{figure}[thbp]
	\centering
		\includegraphics[width=0.85\linewidth]{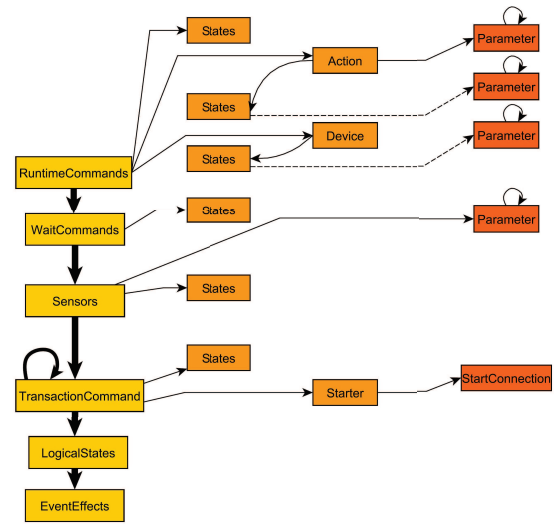}
	\caption{Workflow of the code generator}
	\label{fig:code_generation}
\end{figure}

Once GSRAPID Diagrams have been graphically specified, they can be translated into Java code to be usable inside a RoboticsAPI project.
For this purpose, a code generator can be triggered by the user.
Its main task is to interpret the graphical Command and the values of the respective operator properties.
The challenge here is to correctly process every operator and all inter-operator dependencies (defined by nesting or operator connections).
For this purpose, the instance of the EMF model corresponding to a GSRAPID Diagram is parsed and a Java Abstract Syntax Tree is generated, according to an algorithm whose flow is roughly depicted in Fig.~\ref{fig:code_generation}.

Each entity evaluated by the code generator is checked for contained States, which are translated to Java code first.
Depending on the type of State, its parameters have to be translated before translating the State itself.
Additionally, for TransactionCommands, inner Commands are handled before the TransactionCommand itself.

The translation of LogicalStates is delayed until all other States have been handled. This ensures that all ``precondition'' States have already been translated and the result can be used in the translation of LogicalStates. EventEffects are translated after all (Logical)States habe been processed. At this point during the process, every source and target object for the EventEffect connection is guaranteed to be initialized. The method stubs and member variable declarations are taken care of whenever needed during code generation. The result of the code generation is a Java class with static setter methods for property variables and a static method that instantiates the Command structure as defined with GSRAPID. The implementation of this method checks whether all generated setter methods have been called before, otherwise an error is thrown.

\section{Example}
\label{sec:examples}

To demonstrate the usage and expressiveness of GSRAPID, we present in this section an example Diagram and the steps required to execute the code generated from it. The task that should be realized can be described as follows: Two Lightweight Robots should carry some workpiece cooperatively. During their motion, they pass a container, in which they should drop the workpiece. This task clearly contains some real-time aspects: The motion of both robots must be synchronized in order to achieve a cooperative transport. Furthermore, dropping the workpiece by opening the grippers must be triggered at exactly the right time during movement in order to not miss the container. Both grippers have to be opened exactly at the same time as well, otherwise the workpiece might be stuck too long in one of the grippers and also miss the container. To further increase the complexity, the case that the workpiece actually got stuck during dropping should be handled. This can be done by observing the downward force measured by the robot arms. If the grippers have already been opened and still a significant downward force is measured, the workpiece is assumed to be stuck. In this case, movement of the robots should be stopped immediately.

\begin{figure*}[htbp]
	\centering
		\includegraphics[width=0.95\textwidth]{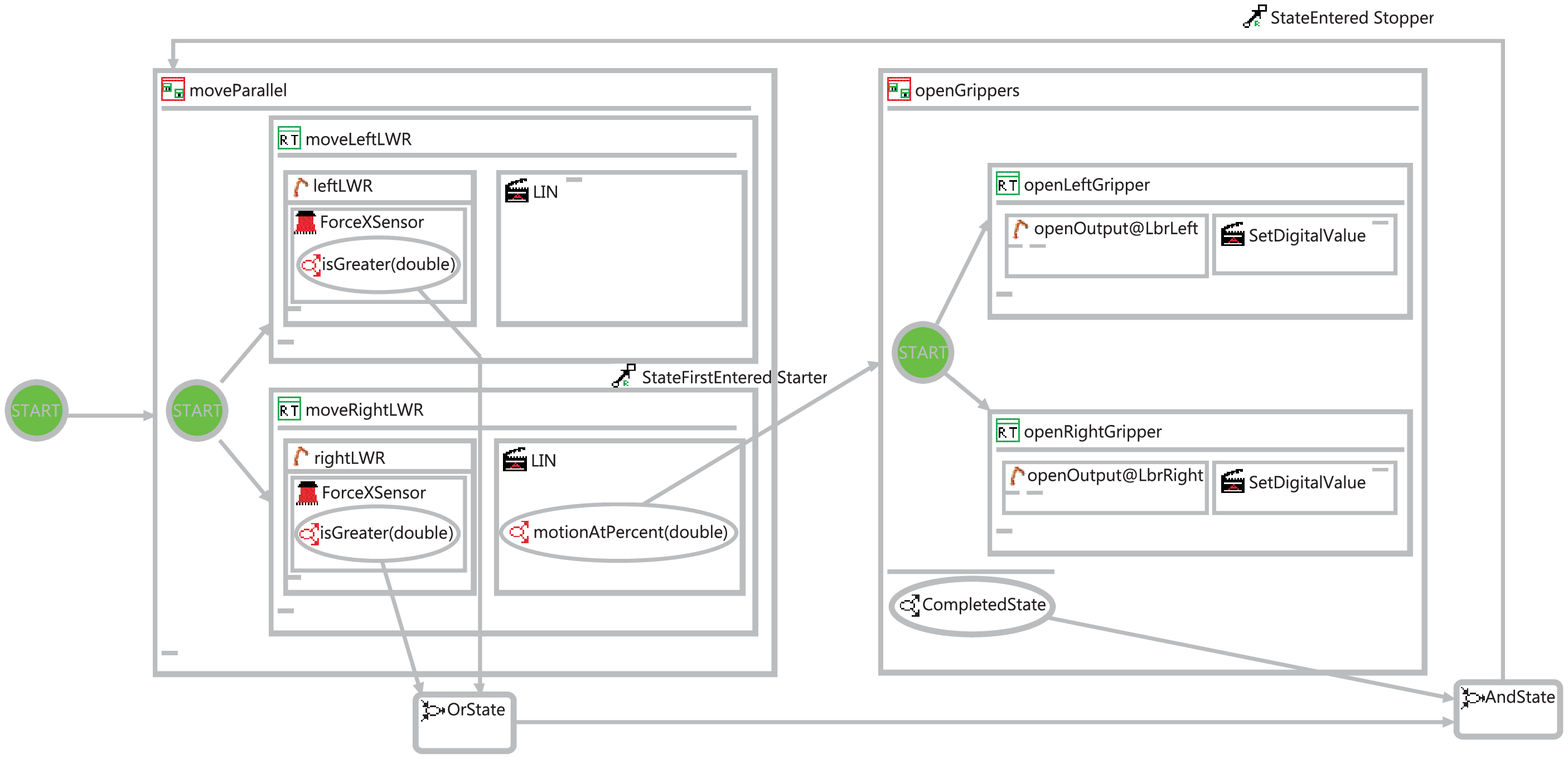}
	\caption{Example Diagram created with the GSRAPID IDE}
	\label{fig:example}
\end{figure*}

Fig.~\ref{fig:example} shows a GSRAPID Diagram that realizes this task by combining all necessary actions in one RoboticsAPI Command. The TransactionCommand in the left part of the Diagram (\emph{moveParallel}) contains two RuntimeCommands, each one controlling one of the robot arms (\emph{leftLWR} and \emph{rightLWR}). Both RuntimeCommands let the robots execute a linear motion (\emph{LIN}). The parameters of these linear motions are defined to be variables, which have to be set before the code generated from the Diagram can be executed (cf.~Listing~\ref{lst:example}). Assuming appropriate parameterization of those LIN Actions, the parallel execution of the RuntimeCommands will result in uniform motion of both robot arms.

The right part of the Diagram models two RuntimeCommands inside a second TransactionCommand called \emph{openGrippers}. Those RuntimeCommands execute \emph{SetDigitalValue} Actions for setting high values on specific field bus outputs. These signals tell the robot grippers connected to those field bus outputs to open. The parallel starting of both RuntimeCommands ensures that the grippers are opened at the same time. The \emph{openGrippers} TransactionCommand is started on the occurrence of a State that is supplied by the LIN Action of one of the robots. This Action offers a method \lstinline{motionAtPercent(p)} that supplies a State indicating that the motion has reached \lstinline{p} percent of its execution time. When a State entity is added to the GSRAPID Diagram and is placed inside the LIN Action entity, the GSRAPID IDE offers this method as one possibility for defining the State (cf.~Fig.~\ref{fig:parameter_handling}).

Handling the case of the workpiece getting stuck is particularly interesting: To detect that the workpiece is stuck, the force sensors of both robot arms are observed. This is modelled in the Diagram by adding Sensors to both \emph{leftLWR} and \emph{rightLWR}. After that, the user has to choose which kind of Sensor to use. The method \lstinline{getForceXSensor()} of the robots' class \lstinline{LWR} is offered to the user as one option (cf.~Fig.~\ref{fig:parameter_handling}). The same mechanism is used to determine appropriate States (defined here by the Sensor's method \lstinline{isGreater(f)}) based on the Sensors' measurements. The States are then combined with an \emph{OrState} so that both measurements are considered. Additionally, the \emph{CompletedState} of the TransactionCommand \emph{openGrippers} is used so that the force measurement is only considered after the grippers have been opened. All States are finally combined by an \emph{AndState}, which then triggers an EventEffect that is stopping the \emph{moveParallel} TransactionCommand.

The code generated from this Diagram consist of about 150 lines. Due to its complexity (cf.~Listing~\ref{lst:commandexample}), writing this code by hand is not an easy job. For other programmers that want to re-use or adapt the code, understanding it is even harder. On the other hand, the semantics of the Command can in most cases be inferred at first glance when looking at the GSRAPID Diagram. Listing~\ref{lst:example} shows the usage of the generated code in a robotics application. As the Diagram was entitled ``Example'', a Java class \lstinline{Example} is generated. This class contains static setter methods for all variables used in the Diagram (in this case, the start and goal poses of the LIN Actions) as well as a static method that instantiates the Command as defined in the Diagram.

\begin{lstlisting}[caption=Using code generated from a GSRAPID Diagram,emph={defineDistanceUnderrunStateSetter,distanceToTargetSensor,setLeftRobotStart, setLeftRobotGoal,setRightRobotStart,setRightRobotGoal,createCommand},emphstyle=\textit,label=lst:example]

// Set parameters under-specified in the Diagram
Example.setLeftRobotStart(Frames.LeftRobotStart);
Example.setLeftRobotGoal(Frames.LeftRobotGoal);

Example.setRightRobotStart(Frames.RightRobotStart);
Example.setRightRobotGoal(Frames.RightRobotGoal);

// At this point Command instantiaton will succeed.
Command exampleCommand = Example.createCommand();

// Finally, Command can be executed.
exampleCommand.execute();
\end{lstlisting}

\section{Conclusion and Future Work}
\label{sec:conclusion}

The GSRAPID language and its IDE promises to be a very useful tool for specifying complex robot commands with the RoboticsAPI. First experiences showed that the process of defining Commands is really accelerated and the resulting diagrams are quickly understandable. However, we have to investigate the applicability of GSRAPID to more complex examples to find out how well the graphical specification scales with the problem complexity. 

The future plans with GSRAPID are twofold: On the one hand, we want to further test GSRAPID with different kinds of hardware and different kinds of tasks to further explore its usefulness as well as potential for improvements. On the other hand, some of the more special Robotics API concepts have not yet been integrated in the GSRAPID language. An example are so called \emph{WrappedActions} that can be used for nesting Actions to e.g. monitor safety constraints of Action implementations. We want to extend GSRAPID in this direction. These extensions will also be interesting to find out the effort required for extending the language and the respective EMF, GMF and GEF models. Until now, our experience is that these technologies are powerful and well designed, though their complexity requires some learning time.

\bibliographystyle{IEEEtran}
\bibliography{paper}

\begin{thebibliography}{10}
\providecommand{\url}[1]{#1}
\csname url@samestyle\endcsname
\providecommand{\newblock}{\relax}
\providecommand{\bibinfo}[2]{#2}
\providecommand{\BIBentrySTDinterwordspacing}{\spaceskip=0pt\relax}
\providecommand{\BIBentryALTinterwordstretchfactor}{4}
\providecommand{\BIBentryALTinterwordspacing}{\spaceskip=\fontdimen2\font plus
\BIBentryALTinterwordstretchfactor\fontdimen3\font minus
  \fontdimen4\font\relax}
\providecommand{\BIBforeignlanguage}[2]{{%
\expandafter\ifx\csname l@#1\endcsname\relax
\typeout{** WARNING: IEEEtran.bst: No hyphenation pattern has been}%
\typeout{** loaded for the language `#1'. Using the pattern for}%
\typeout{** the default language instead.}%
\else
\language=\csname l@#1\endcsname
\fi
#2}}
\providecommand{\BIBdecl}{\relax}
\BIBdecl

\bibitem{Pires2009}
J.~N. Pires, ``New challenges for industrial robotic cell programming,''
  \emph{Industrial Robot}, vol.~36, no.~1, 2009.

\bibitem{Hoffmann2009}
A.~Hoffmann, A.~Angerer, F.~Ortmeier, M.~Vistein, and W.~Reif, ``Hiding
  real-time: A new approach for the software development of industrial
  robots,'' in \emph{Proc. 2009 IEEE/RSJ Intl. Conf. on Intelligent Robots and
  Systems (IROS~2009), St. Louis, Missouri, USA}.\hskip 1em plus 0.5em minus
  0.4em\relax IEEE, Oct. 2009, pp. 2108--2113.

\bibitem{Angerer2010}
A.~Angerer, A.~Hoffmann, A.~Schierl, M.~Vistein, and W.~Reif, ``The {Robotics
  API}: {A}n object-oriented framework for modeling industrial robotics
  applications,'' in \emph{Proc. 2010 IEEE/RSJ Intl. Conf. on Intelligent
  Robots and Systems (IROS~2010), Taipeh, Taiwan}.\hskip 1em plus 0.5em minus
  0.4em\relax IEEE, Oct. 2010, pp. 4036--4041.

\bibitem{Vistein2010}
M.~Vistein, A.~Angerer, A.~Hoffmann, A.~Schierl, and W.~Reif, ``Interfacing
  industrial robots using realtime primitives,'' in \emph{Proc. 2010 IEEE Intl.
  Conf. on Automation and Logistics (ICAL~2010), Hong Kong, China}.\hskip 1em
  plus 0.5em minus 0.4em\relax IEEE, Aug. 2010, pp. 468--473.

\bibitem{Bruyninckx2001}
H.~Bruyninckx, ``Open robot control software: the {OROCOS} project,'' in
  \emph{Proc. 2001 IEEE Intl. Conf. on Robotics and Automation}, Seoul, Korea,
  May 2001, pp. 2523--2528.

\bibitem{Schneider1998}
S.~A. Schneider, V.~W. Chen, G.~Pardo-Castellote, and H.~H. Wang,
  ``{ControlShell}: A software architecture for complex electromechanical
  systems,'' \emph{International Journal of Robotics Research}, vol.~17, no.~4,
  pp. 360--380, 1998.

\bibitem{Borrelly1998}
J.-J. Borrelly, E.~Coste-Mani\`{e}re, B.~Espiau, K.~Kapellos,
  R.~Pissard-Gibollet, D.~Simon, and N.~Turro, ``The {ORCCAD} architecture,''
  \emph{Intl. J. of Robotics Research}, vol.~17, no.~4, pp. 338--359, Apr.
  1998.

\bibitem{Bredenfeld2001}
A.~Bredenfeld and G.~Indiveri, ``{R}obot behavior engineering using
  {DD}-{D}esigner,'' in \emph{Robotics and Automation, 2001. Proceedings 2001
  ICRA. IEEE International Conference on}, vol.~1, 2001, pp. 205 -- 210 vol.1.

\bibitem{MacKenzie1997}
\BIBentryALTinterwordspacing
D.~C. MacKenzie, R.~Arkin, and J.~M. Cameron, ``Multiagent mission
  specification and execution,'' \emph{Autonomous Robots}, vol.~4, pp. 29--52,
  1997, 10.1023/A:1008807102993. [Online]. Available:
  \url{http://dx.doi.org/10.1023/A:1008807102993}
\BIBentrySTDinterwordspacing

\bibitem{Bischoff2002}
R.~Bischoff, A.~Kazi, and M.~Seyfarth, ``The {MORPHA} style guide for
  icon-based programming,'' in \emph{Proc. 11th IEEE International Workshop on
  Robot and Human Interactive Communication}, 2002, pp. 482--487.

\bibitem{Chen2009}
H.~Z. Jing~Chen, Yongzhi~Huang, ``{A} picture or a thousand words? -
  {G}raphic-based robot programming simplifies cell engineering,''
  \emph{{V}ector}, pp. 43--45, April 2009.

\bibitem{Morgan2008}
S.~Morgan, \emph{{P}rogramming {M}icrosoft {R}obotics {S}tudio}, 1st~ed.\hskip
  1em plus 0.5em minus 0.4em\relax Microsoft Press, March 2008, iSBN
  978-0735624320.

\bibitem{Pot2009}
E.~Pot, J.~Monceaux, R.~Gelin, and B.~Maisonnier, ``{C}horeographe: a graphical
  tool for humanoid robot programming,'' in \emph{Proc. 18th IEEE Intl.
  Symposium on Robot and Human Interactive Communication}, 2009, pp. 46--51.

\bibitem{Hoffmann2010a}
A.~Hoffmann, A.~Angerer, A.~Schierl, M.~Vistein, and W.~Reif, ``Towards
  object-oriented software development for industrial robots,'' in \emph{Proc.
  7th Intl. Conf. on Informatics in Control, Automation and Robotics
  (ICINCO~2010), Funchal, Madeira, Portugal}, vol.~2.\hskip 1em plus 0.5em
  minus 0.4em\relax INSTICC Press, Jun. 2010, pp. 437--440.

\bibitem{Schierl2012a}
A.~Schierl, A.~Angerer, A.~Hoffmann, M.~Vistein, and W.~Reif, ``From robot
  commands to real-time robot control - transforming high-level robot commands
  into real-time dataflow graphs,'' in \emph{Proc. 2012 Intl. Conf. on
  Informatics in Control, Automation and Robotics, Rome, Italy}, 2012.

\bibitem{rmi_transparency}
P.~Vign{\'e}ras, ``Transparency and asynchronous method invocation,'' in
  \emph{Proceedings of the 2005 Confederated international conference on On the
  Move to Meaningful Internet Systems - Volume Part I}, ser. OTM'05.\hskip 1em
  plus 0.5em minus 0.4em\relax Berlin, Heidelberg: Springer-Verlag, 2005, pp.
  750--762.

\bibitem{ast}
\BIBentryALTinterwordspacing
T.~Kuhn and O.~Thomann, \emph{Abstract Syntax Tree}, 2006 (accessed august
  2012). [Online]. Available:
  \url{http://www.eclipse.org/articles/article.php?file=Article-JavaCodeManipulation_AST/index.html}
\BIBentrySTDinterwordspacing

\end{thebibliography}

\end{document}